\documentclass[10pt,twocolumn,letterpaper]{article}

\usepackage{cvpr}              % To produce the CAMERA-READY version
\usepackage{graphicx}
\usepackage{amsmath}
\usepackage{amssymb}
\usepackage{booktabs}
\usepackage{multirow}
\usepackage{makecell}
\usepackage{adjustbox}

\makeatletter
\DeclareRobustCommand\onedot{\futurelet\@let@token\@onedot}
\def\@onedot{\ifx\@let@token.\else.\null\fi\xspace}

\def\ie{\emph{i.e}\onedot}

\def\etal{\emph{et al}\onedot}
\makeatother

\usepackage[pagebackref,breaklinks,colorlinks]{hyperref}

\usepackage[capitalize]{cleveref}
\crefname{section}{Sec.}{Secs.}
\Crefname{section}{Section}{Sections}
\Crefname{table}{Table}{Tables}
\crefname{table}{Tab.}{Tabs.}

\def\cvprPaperID{390} % *** Enter the CVPR Paper ID here
\def\confName{CVPR}
\def\confYear{2022}

\begin{document}

%%%%%%%%% TITLE
\title{EfficientNeRF: Efficient Neural Radiance Fields}
%	\tile{EfficientNeRF: Rethinking Sampling in Neural Radiance Fields for }

%Fast and Accurate Novel Scene Synthesis \\ Via Hierarchical Sparse Sampling and NerfTree Representation

\author{
	Tao Hu$^1$ \quad Shu Liu $^2$ \quad  Yilun Chen $^1$ \quad Tiancheng Shen $^1$ \quad Jiaya Jia$^{1,2}$\\
	$^1$ The Chinese University of Hong Kong \qquad
	$^2$ SmartMore\\
	{\tt\small \{taohu,ylchen,tcshen,leojia\}@cse.cuhk.edu.hk , sliu@smartmore.com}
}

\maketitle

%%%%%%%%% ABSTRACT
\begin{abstract}
	Neural Radiance Fields (NeRF) has been wildly applied to various tasks for its high-quality representation of 3D scenes. It takes long per-scene training time and per-image testing time. In this paper, we present EfficientNeRF as an efficient NeRF-based method to represent 3D scene and synthesize novel-view images. Although several ways exist to accelerate the training or testing process, it is still difficult to much reduce time for both phases simultaneously. We analyze the density and weight distribution of the sampled points then propose valid and pivotal sampling at the coarse and fine stage, respectively, to significantly improve sampling efficiency. In addition, we design a novel data structure to cache the whole scene during testing to accelerate the rendering speed. Overall, our method can reduce over 88\% of training time, reach rendering speed of over 200 FPS, while still achieving competitive accuracy. Experiments prove that our method promotes the practicality of NeRF in the real world and enables many applications. The code is available in \href{https://github.com/dvlab-research/EfficientNeRF}{https://github.com/dvlab-research/EfficientNeRF}.
\end{abstract}

%%%%%%%%% BODY TEXT
\section{Introduction}
\label{sec:intro}

Novel View Synthesis (NVS) aims to generate images at new views, given multiple camera-calibrated images. It is an effective line for realizing Visual or Augmented Reality.
With Neural Radiance Fields (NeRF) \cite{NeRF} proposed, NVS tasks \cite{Deformable_NeRF,D_NeRF}, like large-scale or dynamic synthesis \cite{Nerfies,D-NeRF,Animatable-NeRF}, were successfully dealt with in high quality. 
NeRF adopts implicit functions to directly map 3D-point spatial information, in terms of locations and directions, to the attributes of color and densities. To synthesize high-resolution images, NeRF needs to densely sample points over the whole scene, which consumes far more computation than traditional solutions \cite{SRN,NV,LLFF}. For instance, for a scene containing $100$ images with resolution $800 \times 800$, NeRF training time usually takes 1-2 days \cite{NeRF}, and the per-image testing time is around $30$ seconds. These two inefficiencies impede the fast practical applications of NeRF. 

\begin{figure}
	\centering
	\includegraphics[width=1.0\linewidth]{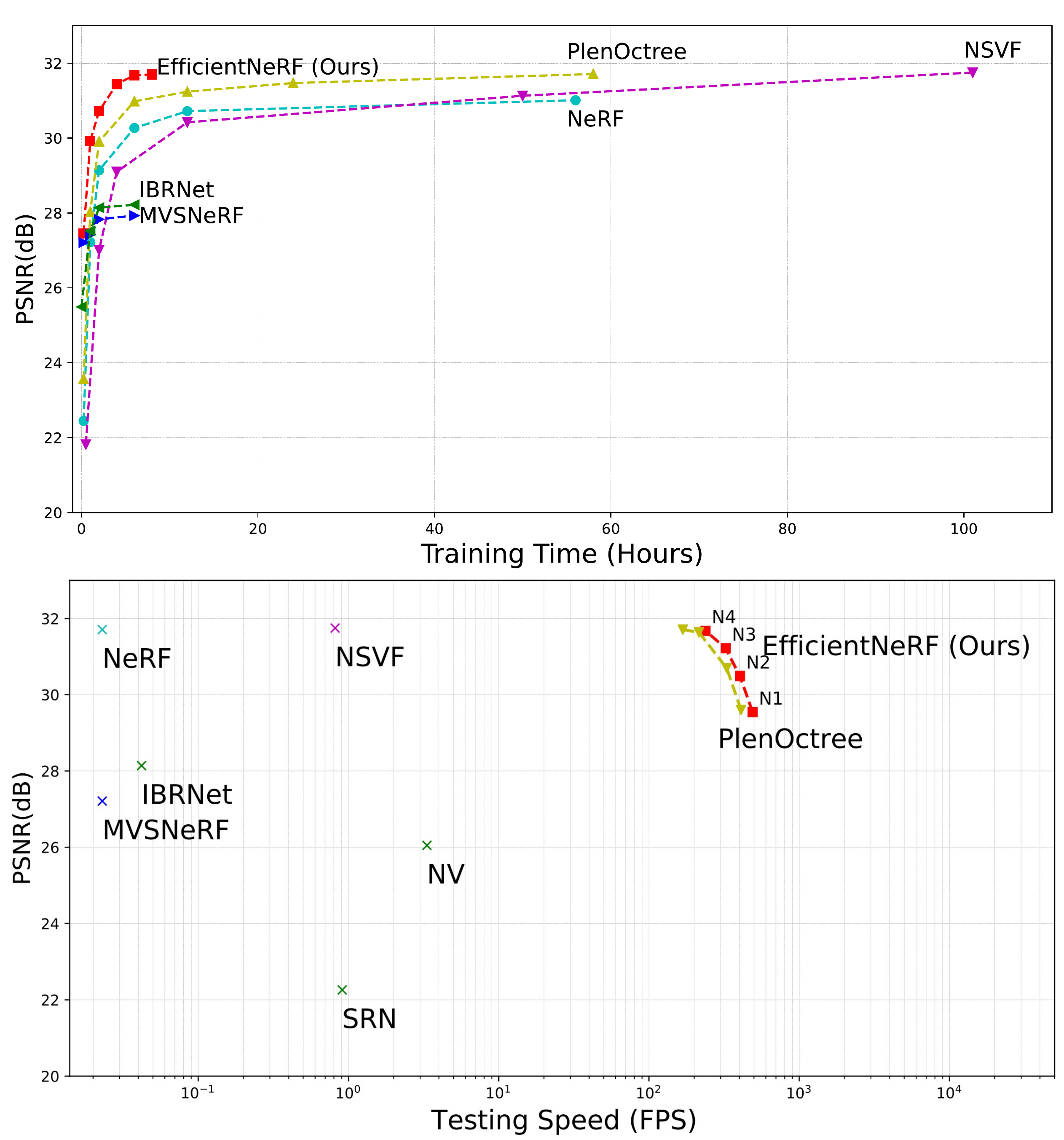}
	\caption{Training and testing efficiency on realistic synthetic dataset \cite{NeRF} on a single GPU. Our EfficientNeRF much improves efficiency in both training and testing phases. }
	\label{fig:time_psnr}
\end{figure}

Recently, methods \cite{NSVF,DONeRF,MVSNeRF,IBRNet,PlenOctree,FastNeRF} were proposed to accelerate either the training process or the testing phase. On the one hand, during testing, NSVF \cite{NSVF} and DONeRF \cite{DONeRF} decrease the number of samples by their generated sparse voxels or predicted depth. FastNeRF \cite{FastNeRF} and PlenOctree \cite{PlenOctree} discretely cache the target scene and synthesize novel-view images by fast query. Although these methods successfully reduce the per-image inference time, their training time is equivalent or even longer, as illustrated in Fig. \ref{fig:time_psnr}. 

On the other hand, during training, methods of \cite{PixelNeRF,MVSNeRF,IBRNet} combine NeRF with image features extracted from ResNet \cite{ResNet} or MVSNet \cite{MVSNet} to construct a generalized model, thus achieving fast training. Nevertheless, as the image prior comes from limited neighboring views, the synthesis accuracy tends to be lower than NeRF \cite{MVSNeRF,IBRNet}. Besides, obtaining features from multi-view images takes more time during testing. There is no work yet to significantly shorten both training and testing time simultaneously. 

In this paper, we present the Efficient Neural Radiance Fields (EfficientNeRF) as the first attempt to accelerate both per-scene training and per-image testing. Apart from obtaining competitive accuracy, the training time can be reduced by more than $88\%$, and the rendering speed is accelerated to over 200 FPS, as illustrated in Fig. \ref{fig:time_psnr}.

The pipeline of original NeRF \cite{NeRF} consists of the coarse and fine stages. During training, the coarse stage obtains the density distribution over the whole scene. It uniformly and densely samples points and calculates corresponding densities by a coarse MLP. However, as will be shown in Table \ref{tab:valid_pivotal_samples}, for common scenes with uniformly sampling, there are only around $10\%$ - $20\%$ of valid samples (in Eq. (\ref{equ:valid_samples})) -- $5\%$ - $10\%$ are pivotal samples (in Eq. (\ref{equ:pivotal_samples})). 

Also, since each point's density is shared by all rays, it is possible to memorize the global density by Voxels. Although NSVF \cite{NSVF} also marks this fact, its solution is to gradually delete invalid voxels, which may cause adverse effects when removal is wrong. Differently, we propose Valid Sampling, which maintains dense voxels and updates density in an online way with momentum. The coarse MLP only infers valid samples whose queried densities are greater than a threshold, thus saving most of the time at the coarse stage. 

For the fine stage, the original NeRF samples more points following previous coarse density distribution. We find that many rays even do not contain any valid and pivotal points because of the empty background. We instead propose Pivotal Sampling for the fine stage that focuses on the nearby area of pivotal samples to efficiently sample points. Our strategy substantially decreases the number of sampled points while achieving comparable accuracy.

During testing, inspired by \cite{PlenOctree} and \cite{FastNeRF} that replace MLP modules by caching the whole scene in voxels, we design a novel tree-based data structure, \ie NerfTree, to more efficiently represent 3D scenes. Our NerfTree only has $2$ layers. The first layer represents the coarse dense voxels extracted from the coarse MLP, and the second layer stores the fine sparse voxels obtained from the fine MLP. The combination of our dense and sparse voxels leads to fast inference speed. 

Our main contributions are the following.
\begin{enumerate}
	\item We propose EfficientNeRF, the first work to significantly accelerate both training and testing of NeRF-based methods while maintaining reasonable accuracy.
	\item We propose Valid Sampling, which constructs dynamic Voxels to accelerate the sampling process at the coarse stage. Also, we propose Pivotal Sampling to accelerate the fine stage. They in total reduce over $88\%$ of computation and training time.
	\item We design a simple and yet efficient data structure, called NerfTree, for NeRF-based methods. It quickly caches and queries 3D scenes, thus improving the rendering speed by $4,000+$ times.
\end{enumerate}

\section{Related Work}
\label{sec:related}
\paragraph{Novel View Synthesis}
NVS is a long-standing problem in computer vision and computer graphics. Voxel grids \cite{Voxel_Coloring,Shape_Carving,NV,MVSM} can achieve real-time synthesis. But they are challenging to represent high-resolution images with large memory consumption. MPI-based methods \cite{Stereo_Matching,soft_3D,Stereo_magnification,View_Extrapolation,LLFF,Nex} can synthesize high-resolution images. They first synthesize multiple depth-wise images and then fuse them to the target views by $\alpha$-compositing \cite{Alpha_compositing}. Large-view synthesis \cite{IBRNet} is necessary. 

% \vspace{-0.1in}
\paragraph{NeRF-based Applications}
NeRF \cite{NeRF} resolves the resolution and memory issues, and can be easily expanded to various applications. Nerfies \cite{Nerfies}, NSVF \cite{NSVF}, and D-NeRF \cite{D-NeRF} implicitly learn 3D spatial deformation functions for dynamic scenes where objects are moving in different frames. Neural Actor \cite{Neural_Actor} and Animatable-NeRF \cite{Animatable-NeRF} also adopt similar functions to synthesize human body with novel poses. GRAF \cite{GRAF}, pi-GAN \cite{pi-GAN}, and GIRAFFE \cite{GIRAFFE} treat NeRF as a generator in GAN \cite{GAN} and generate geometrically controllable images. Recently, StyleNeRF \cite{StyleNeRF} succeeds in generating images at $1$K resolution, which encourages development of NeRF generator. 

\vspace{-0.1in}
\paragraph{Training Acceleration}
Training of NeRF \cite{NeRF} and its variants usually takes 1 to 2 days \cite{NeRF,PlenOctree,NSVF}, which limits efficiency-critical applications. Yu \etal proposed PixelNeRF \cite{PixelNeRF} that introduces image features from ResNet \cite{ResNet} and skips training in novel scenes. But its synthesis accuracy reduces \cite{MVSNeRF}. Wang \etal proposed IBRNet \cite{IBRNet} that integrates multi-view features in the weighted sum, thus improving accuracy. Chen \etal proposed MVSNeRF \cite{MVSNeRF},
which employs MVSNet \cite{MVSNet} to provide a feature volume for NeRF. It can synthesize high-quality images within 15-minute finetuning. However, the testing time of the above methods is as long as the original NeRF.

\begin{figure*}[t]
	\centering
	\includegraphics[width=0.999\linewidth]{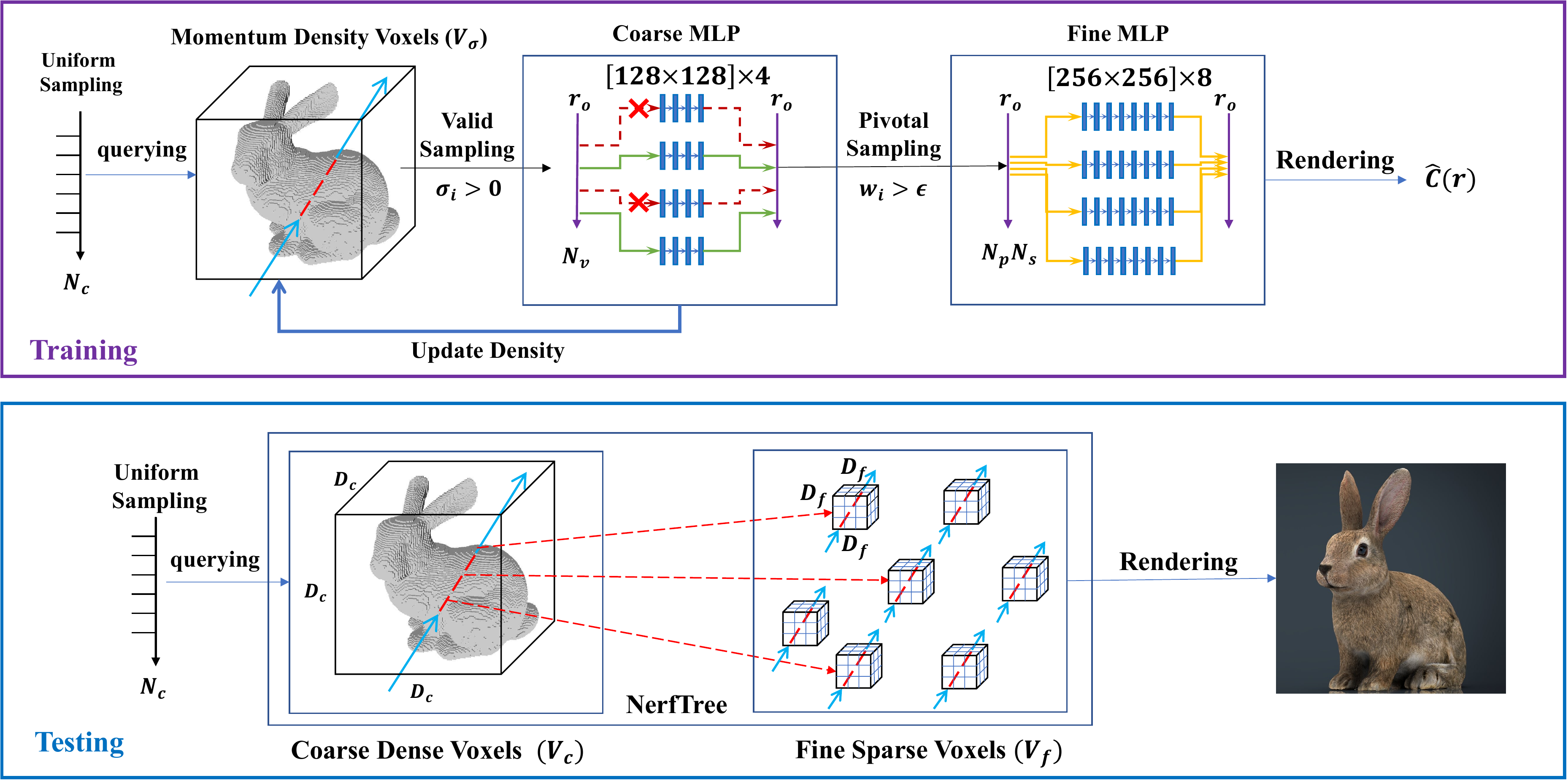}
	\caption{Overview of our proposed EfficientNeRF.  \textbf{Training}: we first uniformly sample $N_c$ points along each ray $\textbf{r}$, and query the density from the Momentum Density Voxels $V_\sigma$. We calculate its coarse density for the valid samples whose density $\sigma_i > 0$, obtain weight to calculate the final ray color, and update $V_\sigma$ by the coarse density. The pivotal samples with weights $w_i > \epsilon$ are taken care of. $N_s$ nearby samples are linearly sampled along ray $\textbf{r}$ at higher resolutions. Finally, we infer the fine density and color parameters by the fine MLP and predict the ray color by volume rendering. $\textbf{Testing}$: The Coarse Dense Voxels and Fine Sparse Voxels are respectively extracted from coarse and fine MLPs. The densities and colors are obtained by voxels query rather than MLPs. 		
	}
	\label{fig:overview}
\end{figure*}

% \vspace{-0.1in}
\paragraph{Testing Acceleration}
To accelerate per-image inference, NSVF \cite{NSVF} gives a hybrid scene representation that combines NeRF with sparse voxels structure. The generated sparse voxels guide and reduce sampling. It improves the inference speed to around $1$ FPS. 
KiloNeRF \cite{KiloNeRF} reduced the inference time by adopting around 1,000 tiny MLPs, where each MLP takes care of a specific 3D area. The running speed is over 10 FPS. PlenOctree \cite{PlenOctree} and FastNeRF \cite{FastNeRF} achieved inference speed over 168 FPS and 200 FPS respectively by caching the whole 3D scenes. We note that their training is still heavy. In contrast, our EfficientNeRF achieves faster per-image inference speed along with far less training time.

\section{Our Approach}
\label{sec:approach}
Given $M$ images ${I_m (m=1,2, ..., M)}$ with calibrated cameras parameters in multiple views of a scene, we aim to achieve accurate 3D scene representation and novel image synthesis regarding both fast training and testing. To begin with, we review the basic idea and pipeline of NeRF \cite{NeRF}.  Then, we introduce our efficient strategies during training, including lightweight MLP, valid sampling at the coarse stage, and pivotal sampling at the fine stage. Finally, we represent the whole scene by our proposed NerfTree during testing to reach hundreds of FPS. 

\begin{table*}[t]
	\begin{adjustbox}{width=1\linewidth,center}
		\begin{tabular}{c|c|c|c|c|c|c|c|c|c}
			\hline
			\textbf{Scene} & Chair & Drums & Ficus & Hotdog & Lego & Materials & Mic & Ship & \textbf{Mean} \\ \hline
			\textbf{Valid Samples (V, \%)} & 9.58 \% & 7.00
			\% & 3.85 \% &  9.35 \% &  15.43 \%  & 19.47 \%   & 8.44 \% & 11.32 \% &  \textbf{10.56} \% \\
			\textbf{Pivotal Samples (P, \%)} & 3.79 \% & 2.25\% &  1.68 \% &  3.59 \% & 5.81  \%  &7.42 \%  & 3.14 \% & 4.62 \% &  \textbf{4.04 }\% \\ \hline
			\textbf{Scene} & Fern & Flower & Fortress & Horns & Leaves & Orchids & Room & Trex & \textbf{Mean} \\ \hline
			\textbf{Valid Samples (V, \%)} & 24.28 \% & 13.68 \%  &23.45 \%  &  21.34 \%& 15.09\% &  19.74 \% &30.62 \%  &18.27 \%   & \textbf{20.81}\% \\ 
			\textbf{Pivotal Samples (P, \%)} & 15.63 \% &7.49 \%   &4.48 \%  &  10.45 \% & 8.49\% &   9.43 \% &15.23 \%  &7.89 \%   &  \textbf{9.89}\% \\ \hline
		\end{tabular}
	\end{adjustbox}
	\caption{Proportions of valid and pivotal samples on the Realistic Synthetic dataset \cite{NeRF} and the Real Forward-Facing dataset \cite{LLFF}. }
	\label{tab:valid_pivotal_samples}
\end{table*}

\subsection{Background: 
	Neural Radiance Fields}
\label{sec:nerf}
NeRF \cite{NeRF} is a new representation to 3D scenes. Different from 3D mesh, point clouds, and voxels, it introduces implicit functions to model scenes while adopting volume rendering to synthesize images. Compared with voxels-based representation, NeRF overcomes the limitation of resolution and storage to synthesize high-quality results.

\vspace{-0.1in}
\paragraph{Implicit Function} NeRF employs implicit functions to inference the sampled points' 4D attributes when inputting 5D spatial information, formulated as
\begin{equation}
	(r,g,b,\sigma) = f(x,y,z,\theta,\phi),
	\label{equ:implicit_function}
\end{equation}
where $\textbf{x}=(x,y,z)$ and $\textbf{d}=(\theta,\phi)$ denote the point location and direction in the world coordinate. The color and density attributes are respectively represented by $\textbf{c}=(r,g,b)$ and $\sigma$. $f$ is a mapping function, usually implemented by a MLP network.

\vspace{-0.1in}
\paragraph{Volume Rendering} For each pixel in the synthesized image, to calculate its color, NeRF first samples $N$ points $\textbf{x}_i (i=1,2,...,N)$ along ray $\textbf{r}$. It then calculates corresponding density $\sigma_i$ and color $\textbf{c}_i$ by Eq. (\ref{equ:implicit_function}). The final predicted color $\hat{C}(r)$  is rendered by $\alpha$-compositing \cite{Alpha_compositing} as 
\begin{align}
	\hat{C}(\textbf{r}) &= \sum_{i=1}^N w_i c_i, \\ 
	w_i &= T_i \alpha_i, \nonumber\\
	T_i &= exp(-\sum_{j=1}^{i-1}\sigma_j\delta_j), \nonumber\\
	\alpha_i &= 1 - exp(-\sigma_i\delta_i), \nonumber
	\label{equ:alpha_compositing}
\end{align}
where $\delta_i$ denotes interval of samples along ray $\textbf{r}$.

\vspace{-0.1in}
\paragraph{Training Objective} The training objective $\mathcal{L}$ of NeRF is the mean square error between each ground-truth pixel color $C(\textbf{r})$ and the rendering color $\hat{C}(\textbf{r})$ as
\begin{equation}
	\mathcal{L} = \sum_{\textbf{r} \in \mathcal{R}} \parallel  C(\textbf{r}) - \hat{C}(\textbf{r}) \parallel_2^2,
\end{equation}
where $\mathcal{R}$ is the set of all rays shooting from the camera center to image pixels.

\subsection{Network}
\label{subsec:mlp}
The original NeRF \cite{NeRF} adopts a coarse-to-fine pipeline to represent scenes. There are two Multi-Layer Perceptrons (MLPs) in the model with the same network size, while respectively operate the coarse and fine stages. We call them coarse and fine MLPs. Since the coarse MLP mainly infers a coarse density distribution, original coarse MLP is redundant and reducible. 

For the sake of simplicity, we directly decrease both the depth and width of coarse MLP by half and keep the consistency of the fine MLP, as illustrated in Fig. \ref{fig:overview}. According to the experimental results in Table \ref{tab:training_contribution}, our lightweight coarse MLP almost does not weaken performance while improving the overall inference speed. Taking advantage of our lightweight coarse MLP, we increase $N_c$ to improve the synthesized quality. 

Different from the original NeRF that employs an implicit mapping from direction to color, we adopt the Spherical Harmonics model in PlenOctree \cite{PlenOctree} to explicitly predict color parameters by the MLP network. It not only improves the accuracy but also is beneficial for offline caching during testing.  

\subsection{Valid Sampling at the Coarse Stage}
\label{sec:sampling}

%\begin{figure*}[t]
	%	\centering
	%	\subfloat[A table right of a figure]{
		%		\begin{adjustbox}{width=1.0\columnwidth,left}
			%		\begin{tabular}{c|c|c|c|c|c|c|c|c|c}
				%				\hline
				%				\textbf{Scene} & Chair & Drums & Ficus & Hotdog & Lego & Materials & Mic & Ship & \textbf{Mean} \\ \hline
				%				\textbf{Valid Samples (V, \%)} &  &  &  &  &  23.43 &  &  &  &  \\
				%				\textbf{Pivotal Samples (P, \%)} &  &  &  &  &  7.81 &  &  &  &  \\ \hline
				%				\textbf{Scene} & Fern & Flower & Fortress & Horns & Leaves & Orchids & Room & Trex & \textbf{Mean} \\ \hline
				%				\textbf{Valid Samples (V, \%)} &  &  &  &  &  &  &  &  &  \\ 
				%				\textbf{Pivotal Samples (P, \%)} &  &  &  &  &  &  &  &  &  \\ \hline
				%			\end{tabular}
			%		\end{adjustbox}
		%	}
	%	\subfloat[A figure left of a table]{
		%	\includegraphics[width=\columnwidth]{figs/distribution.pdf}
		%	}
	%	\caption{A figure and a table, side-by-side}
	%\end{figure*}

\begin{figure}[t]
	\centering
	\includegraphics[width=1\linewidth]{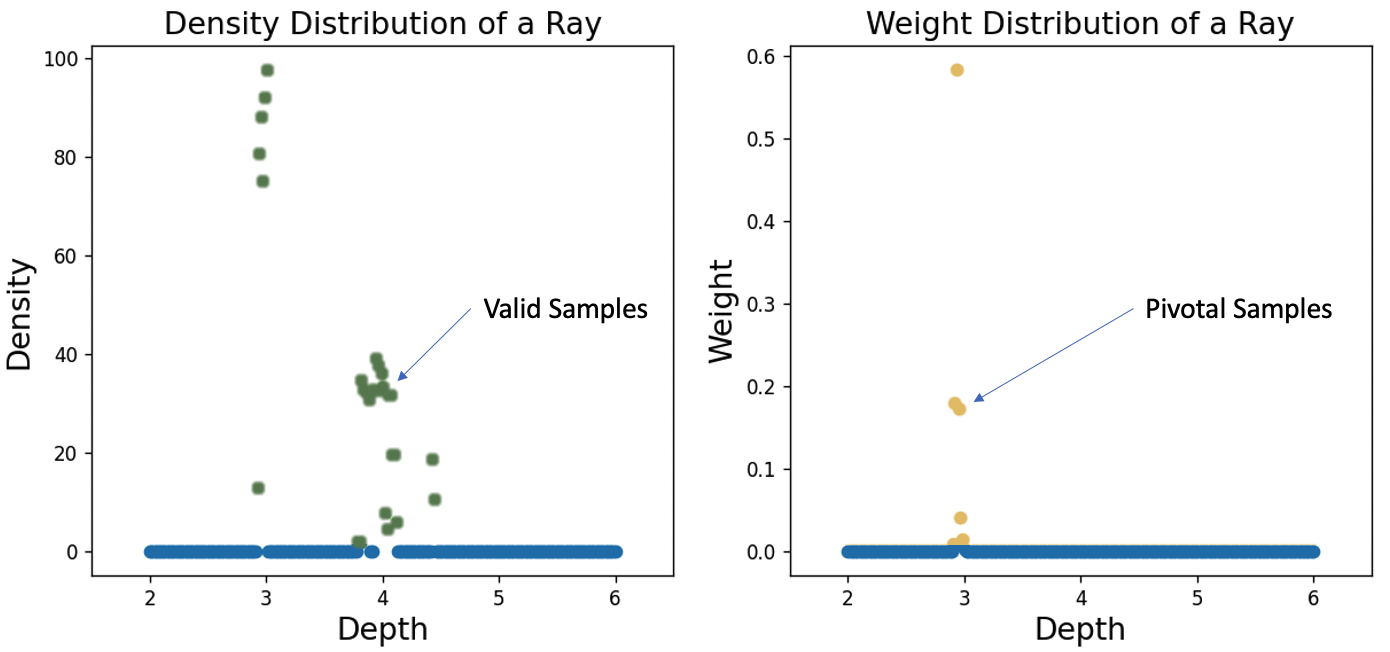}
	\caption{Density and weight distributions of a typical ray for NeRF-based methods. Green and yellow points indicate valid and pivotal samples, respectively. }
	\label{fig:raydistri}
\end{figure}

\paragraph{Valid Samples} We define the point with location $\textbf{x}_v$ with density  $\sigma_v > 0 $ as a valid sample, as shown in Fig. \ref{fig:raydistri}. 
For the $N$ points along ray $\textbf{r}$, suppose a point with location $\textbf{x}_i$ has density $\sigma_i=0$. Because $T_i = exp(-\sum_{j=1}^{i-1}\sigma_j\delta_j)$ belongs to the range of $[0, 1]$ and $
\alpha_i = 1 - exp(-\sigma_i\delta_i) = 0$, we calculate its contribution $w_i$ to the ray color $\hat{C}(r)$ by 
\begin{equation}
	w_i = T_i \cdot \alpha_i  = 0.
\end{equation}
It means the point is an invalid sample and makes no difference to the final rendering result. Therefore, it can be skipped once we know the locations. The proportion of valid samples is represented as 
\begin{equation}
	V = \frac{N_v}{N_c}.
	\label{equ:valid_samples}
\end{equation}

We measure the percentage of area by trained NeRF that is valid over common scenes and show the numbers in Table \ref{tab:valid_pivotal_samples}. It is surprising to note that only a small portion (around $10\%$ - $20\%$) of the samples are valid. From this analysis, we conclude that it is feasible and necessary to adopt sparse and valid sampling to achieve efficient scene representation.

\vspace{-0.1in}
\paragraph{Momentum Density Voxels}
For a specific scene, the density of any world coordinate $\textbf{x}\in \mathbb{R}^{3}$ is shared by all rays. Thus, we construct momentum density voxels $V_\sigma$ with resolution $D \times D \times D$ to memorize the latest global value of density over the target scene during training. 

\vspace{-0.1in}
\paragraph{Initialization}
Since each point's density $\sigma \ge 0$, we initialize the default density value in $V_\sigma$ as a positive number $\varepsilon$. It means that all points in $V_\sigma$ are valid samples in the beginning. 

\vspace{-0.1in}
\paragraph{Update}
For a sampled point with location $\textbf{x} \in \mathbb{R}^3$, we infer its coarse density $\sigma_c(\textbf{x})$  by the coarse MLP. Then we update the density voxels $V_\sigma$ by  $\sigma_c(\textbf{x})$. We add a momentum to stabilize the values. Specifically, we first transfer $\textbf{x}$ to 3D Voxels index $\textbf{i}\in \mathbb{R}^3$ as
\begin{align}
	\textbf{i} &= \dfrac{\textbf{x} - \textbf{x}_{min}}{\textbf{x}_{max} - \textbf{x}_{min}} \cdot D,
	\label{equ:coordinate}
\end{align}
where $\textbf{x}_{min}, \textbf{x}_{max}\in \mathbb{R}^3$ denote the minimal and maximal world coordinate borders of the scene. 

Next, for every training iteration, we update the global density $\sigma$ at index $\textbf{i}$ of $V_\sigma$ through
\begin{equation}
	V_\sigma[\textbf{i}] \leftarrow (1 - \beta) \cdot V_\sigma[\textbf{i}] + \beta\cdot \sigma_c(\textbf{x}).
	\label{equ:momentum}
\end{equation}
Where $\beta\in[0, 1]$ controls the updating rate.

Our momentum density Voxels $V_\sigma$ reflect the latest density distribution over the whole scene. Thus, we directly obtain the density attribute at coordinate $x$ through query rather than calculating through a MLP module. It primarily reduces the inference time and is utilized to guide a dynamic sampling process. 

\vspace{-0.1in}
\paragraph{Valid Sampling}
During training, for each ray $\textbf{r}$ whose starting point is $r_o \in \mathbb{R}^3$ and normalized direction is $r_d \in \mathbb{R}^3$, the original NeRF adopts a uniform sampling strategy to obtain the sampled points as
\begin{equation}
	\textbf{x}_i = r_o + i \delta_c r_d ,
	\label{equ:coarse_sampling}
\end{equation}
where $i\in Z$ and $\forall  i \in [1, N_c] $.  $\delta_c$ is the interval between the nearest coarse sampled points along ray $\textbf{r}$. 

We propose Valid Sampling to pay attention to valid samples. Specifically, instead of directly inferring all these samples, we first query the latest density from $V_\sigma$, and only input $\textbf{x}_i$ with global densities
\begin{equation}
	V_\sigma[\textbf{i}] > 0
	\label{equ:threashold}
\end{equation}
to the coarse MLP.

Inferring a single point by a coarse MLP takes times $T_m$, and querying a single point from voxels takes $T_q$.
For all sampled points along ray $\textbf{r}$, predicting their densities through a coarse MLP consumes time $N_c T_m$. Our method takes time $(N_vT_m + (N_c-N_v)T_q)$. Considering time of voxels query $T_q \ll T_m$ \cite{PlenOctree,FastNeRF}, we calculate the acceleration ratio $A_c$ of the coarse stage by 
\begin{equation}
	A_c = \frac{N_cT_m}{N_vT_m + (N_c-N_v)T_q} \approx \frac{N_c}{N_v} = \dfrac{1}{V}.
\end{equation}
As illustrated in Table \ref{tab:valid_pivotal_samples}, if the proportion of valid samples $V=10 \%$, the coarse stage can be accelerated by $10$ times in theory.

\subsection{Pivotal Sampling at the Fine Stage}
During the fine stage, 3D points should be sampled in higher resolution for better quality. The original NeRF \cite{NeRF} first samples $N_f$ points along each ray $\textbf{r}$ that follows the coarse weight distribution. It then predicts densities and colors by the fine MLP. Since the number of points at the fine stages is usually $2$ times of $N_c$, it requires more computation during running time. To achieve efficient sampling at the fine stage, we propose a Pivotal Sampling strategy.

\vspace{-0.1in}
\paragraph{Pivotal Samples} We define the point with location  $\textbf{x}_p$ whose weight $w_p > \epsilon$ as a pivotal sample, where $\epsilon$ is a tiny threashold, as illustrated in Fig. \ref{fig:raydistri}.

\vspace{-0.1in}
\paragraph{Pivotal Sampling} $w_i$ represents the contribution of $\textbf{x}_i$ to the ray $\textbf{r}$'s color. The nearby area of the pivotal samples is focused to infer more detailed densities and colors. Apart from $x_p$, we uniformly sample $N_s$ points near $\textbf{x}_p$ along each ray $\textbf{r}$ as
\begin{equation}
	\textbf{x}_{p,j} = \textbf{x}_p + j \delta_f r_d,
	\label{equ:pivotal_sampling}
\end{equation} 
where $	j \in Z$ and $\forall j \in [-\frac{N_s}{2},\frac{N_s}{2}]$.  $\delta_f$ is the interval at the fine stage.  Suppose there are $N_p$ pivotal points, the proportion of the pivotal samples can be represented by 
\begin{equation}
	P = \frac{N_p}{N_c}.
	\label{equ:pivotal_samples}
\end{equation}
Similar to the coarse stage, we calculate the acceleration ratio $A_f$ of the fine stage as
\begin{equation}
	A_f = \dfrac{N_f}{N_pN_s}  = \dfrac{2N_c}{N_pN_s} = \dfrac{2}{PN_s}.
	\label{equ:fine_accelerate}
\end{equation}
In our experiments with results listed in Table \ref{tab:valid_pivotal_samples}, if $N_s=5$ and $P=5\%$, our pivotal sampling strategy can accelerate the fine stage by $8$ times.

\subsection{Represent Scene by NerfTree}
Although the training time has been significantly shortened through our valid and pivotal sampling, the system is still constrained by the inference time of MLP during testing. Inspired by \cite{PlenOctree,FastNeRF} that cache the target scene in Voxels or Octrees, we design an efficient tree-based data structure, called NerfTree, for NeRF-based methods to accelerate the inference speed. NerfTree can store the whole scene offline, thus eliminating the coarse and fine MLP. 

Different from the dense Voxels and Octrees in PlenOctree \cite{PlenOctree}, our NerfTree $T=\{V_c, V_f\}$ is a 2-depth tree. The first depth caches the coarse dense Voxels $V_c$, and the second depth caches the fine sparse Voxels $V_f$, as illustrated in Fig. \ref{fig:overview}. $V_c \in \mathbb{R}^{D_c\times D_c \times D_c}$ only contains density attribute, which is extracted by inferring the density values of every voxel grid by the coarse MLP.

\begin{figure}[t]
	\centering
	\includegraphics[width=1.0\linewidth]{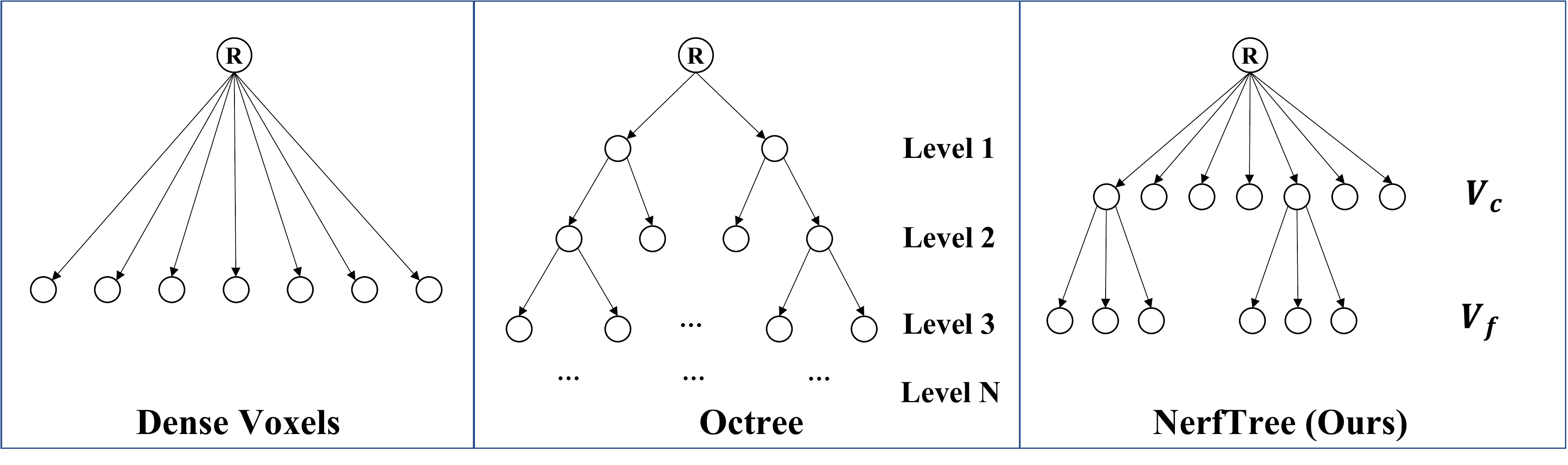}
	\caption{2D graph representation of different 3D data structures.  Left: Dense Voxels. Middle: Octrees. Right: NerfTree (Ours). }
	\label{fig:graph}
\end{figure}

For the fine sparse voxels $V_f \in \mathbb{R}^{N_V \times D_f^3}$, the first dimension $N_V$ represents the number of all valid samples, and the second dimension represents local voxels with size $D_f\times D_f\times D_f$. Each voxel in $V_f$ stores the density and color parameters inferred from the fine MLP. As illustrated in Fig. \ref{fig:graph}, between these three representations, dense voxels only have one depth layer, thus achieving the minimal access time and maximum storage. Octree has the opposite characteristic of Voxels. Our NerfTree combines the advantages of both Voxels and Octrees. Thus it can be accessed at a fast speed while not consuming much storage.

\section{Experiments}
\label{sec:exp}
%In this section, we conduct experiments to quantitatively and qualitatively evaluate the effectiveness of our EfficientNeRF in both speed and accuracy. 

\begin{table*}[h]
	\begin{adjustbox}{width=2.0\columnwidth,center}
		\begin{tabular}{c|c|c|c|c|c|c|c|c|c|c}
			\hline
			\multirow{2}{*}{Method} & \multicolumn{5}{c|}{Realistic Synthetic \cite{NeRF}} & \multicolumn{5}{c}{Real Forward Facing \cite{LLFF,NeRF}} \\ \cline{2-11} 
			& PSNR($ \uparrow $) & SSIM ($ \uparrow $) & LPIPS ($\downarrow$) & \makecell{Training Time \\ (Hours, $\downarrow$)} &\makecell{Rendering Speed \\ (FPS, $\uparrow$)} & PSNR ($\uparrow$) & SSIM ($\downarrow$) & LPIPS ($\downarrow$) & \makecell{Training Time \\ (Hours, $\downarrow$)} & \makecell{Rendering Speed \\ (FPS, $\uparrow$)} \\ \hline
			SRN \cite{SRN} & 22.26 & 0.846 & 0.170 & - & 0.909 &  22.84&  0.668 &  0.378& - &  - \\ 
			NV \cite{NV} & 26.05 & 0.893 & 0.160 & - & 3.330 & - & - & - & - & 3.052 \\
			MVSNeRF \cite{MVSNeRF} & 27.21 & 0.945 & 0.227 & \textbf{0.25} & 0.020 & 26.25 &  \textbf{0.907} &  0.139 &  0.25 &  0.016\\ 
			IBRNet \cite{IBRNet} & 28.14 & 0.942 & 0.072 & \textbf{2.0 }& 0.042 &  26.73 & 0.851 & 0.175 & \textbf{2.0 }& 0.036 \\ 
			%			\textbf{EfficientNeRF (15-min)} & 28.73 & 0.946 & 0.075 & 0.25 & 238.46 &  &  &  &  &  \\ 
			\hline
			NeRF \cite{NeRF} & 31.01 & 0.947 & 0.081 & 56 & 0.023 &  26.50& 0.811 &  0.250 &  20 & 0.018 \\ 
			NSVF \cite{NSVF} & \textbf{31.75} & 0.953 & 0.047 & 100+ & 0.815 & - & - & - & - & 0.758\\ 
			AutoInt \cite{AutoInt} & 25.55 & 0.911 & 0.170 & - & 0.380 & 24.13 & 0.820 & 0.176 & - & -\\ 
			KiloNeRF \cite{AutoInt} & 31.00 & 0.950 & \textbf{0.030} & 25+ &  10.64 & - & - & - & - &  -\\ 
			FastNeRF \cite{FastNeRF} & - & - & - & - & $\sim$200  & 26.04 & 0.856 & \textbf{0.085} & - & $\sim$200  \\ 
			Nex \cite{Nex} & - & - & - & - & - &\textbf{27.26} & 0.904 & 0.178 & 18+ & \textbf{300}  \\ 
			PlenOctree \cite{PlenOctree} & \textbf{31.71} & \textbf{0.958} & 0.053 & 58 & \textbf{167.68} &  -&  -&  -&  - & - \\ 
			\hline
			\textbf{EfficientNeRF}  & 31.68 & \textbf{0.954}& \textbf{0.028} & 6 & \textbf{238.46} & \textbf{27.39 } & \textbf{0.912} & \textbf{0.082}&\textbf{4} & \textbf{218.83}  \\ \hline
		\end{tabular}
	\end{adjustbox}
	\caption{Accuracy and time comparison on the Realistic Synthetic \cite{NeRF} and the Real Forward-Facing \cite{NeRF,LLFF} datasets. Ours achieves comparable PSNR/SSIM/LPIPS accuracy with state-of-the-art methods, while showing promising acceleration in both training and testing phases.}
	\label{tab:quantative_comparison}
\end{table*}

\subsection{Experimental Setting}
\paragraph{Datasets}
We first introduce the common high-resolution Novel View Synthesis datasets, including the Realistic Synthetic dataset \cite{NeRF} and the Real Forward-Facing dataset \cite{LLFF}. The Realistic Synthetic dataset \cite{NeRF} contains 8 synthetic scenes. Each scene contains 100 training images and 200 testing images, all at $800\times 800$ resolution. The Real Forward-Facing dataset \cite{LLFF} consists of 8 complex and real-world scenes, each has 20 to 62 images at $1,008\times 756$ resolution. We follow the same training and testing dataset split as the original NeRF \cite{NeRF}. 

\vspace{-0.1in}
\paragraph{Metrics}  We evaluate the accuracy of synthesized images via metrics including PSNR / SSIM (the higher the better), and LPIPS \cite{LPIPS} (the lower the better), following recent methods \cite{NeRF,NSVF,PlenOctree,NV}. Moreover, we measure the training speed by total training time in terms of hour and the rendering speed by Frame Per Second (FPS). For fair comparison, we re-train their source code on the same machine and skip the evaluation time during training.

\vspace{-0.1in}
\paragraph{Implementation Details}
During training, for the Momentum Density Voxels $V_\sigma$, its resolution is set as $384\times 384\times 384$, the initial density $\varepsilon=10.0$, and the updating rate $\beta=0.1$. The degree of Spherical Harmonics is $3$, which means the output dimension of MLPs is $49$ \cite{PlenOctree}.The pivotal threshold $\epsilon$ is $1\times10^{-4}$. The learning rate is initialized to $ 5\times 10^{-4}$ with Adam \cite{Adam} optimizer and exponentially decays by $0.1$ for every $500$K iteration when the batch size is $1024$. Finally, the total number of training iteration is $1 \times10^6$.  

During testing, we set $D_c=384$ and $D_f=4$, which means the coarse voxels have resolution $384\times 384 \times 384$, and local fine sparse voxels have resolution $4 \times 4 \times 4$. The quantification from MLP's continuous coordinates to discrete ones usually weakens the performance \cite{PlenOctree}. We avoid it by directly inputting the converted discrete coordinate to the MLP during training or using linear interpolation. 
All our experiments are performed on a server with one RTX-3090 GPU. Please refer to the supplementary material for more experimental results. 

\subsection{Quantitative Comparison}
We compare the proposed EfficientNeRF with state-of-the-art methods \cite{SRN,NV,NeRF,NSVF,AutoInt,PlenOctree}  in terms of both accuracy and speed. Results are listed in Table \ref{tab:quantative_comparison}. Depending on whether image prior is introduced or not, state-of-the-art methods can be divided into two groups. The first \cite{SRN,NV,MVSNeRF,IBRNet} is based on image prior. These methods can fine-tune a novel scene in short time while sacrificing testing accuracy. The second group \cite{NSVF,AutoInt,PlenOctree} is by training from scratch. They were designed to improve the rendering speed. However, the training time is found even longer than that of the original NeRF \cite{NeRF}.  

In contrast, our EfficientNeRF achieves competitive accuracy on both datasets and demonstrates notable advantage in terms of training and testing efficiency. As shown in Fig. \ref{fig:time_psnr} and Table \ref{tab:quantative_comparison}, even though our method does not introduce image prior, it still outperforms previous fast finetuning method \ie, MVSNeRF \cite{MVSNeRF}, when training for 15 minutes or longer. In addition, our proposed NerfTree quickly queries the 3D attributes at the target locations, which contributes to our final 238.46 FPS during testing. 

In summary, our EfficientNeRF adopts efficient strategies, including lightweight MLP, valid sampling, and pivotal sampling, thus accelerating both the training and testing while maintaining comparable accuracy.

\vspace{-0.1in}
\paragraph{Trade-off between Accuracy and Speed}
To balance the synthesized accuracy and inference speed, we provide four versions of EfficientNeRF ($N1 $-$ N4$) according to the number of coarse and fine sampling parameters $N_c$ and $N_s$ in Table \ref{tab:trade-off} and plot the PSNR-Speed curves in Fig. \ref{fig:time_psnr}. Our work achieves a better rendering speed than other state-of-the-art methods like PlenOctree \cite{PlenOctree}.
%30.07, 30.49, 31.22, 31.68
%fps = [523, 447, 324, 238]
\begin{table}[h]
	\begin{adjustbox}{width=1.0 \columnwidth,center}
		\begin{tabular}{l|c|cc|c|c}
			\hline
			\multirow{2}{*}{} & \multirow{2}{*}{Version} & \multicolumn{2}{c|}{\# Sampling} & \multirow{2}{*}{PSNR ($\uparrow$)} & \multirow{2}{*}{{Rendering Speed (FPS, $\uparrow$)}} \\ \cline{3-4}
			&  & \multicolumn{1}{c|}{Coarse} & {Fine} &  &  \\ \hline
			\multirow{4}{*}{EfficientNeRF} & $ N1 $ & \multicolumn{1}{c|}{64} & 2 & 29.54 & {493.62} \\ \cline{2-6} 
			& $ N2 $ & \multicolumn{1}{c|}{64} & 3 & 30.49 & 403.28 \\ \cline{2-6} 
			& $ N3 $ & \multicolumn{1}{c|}{96} & 4 & 31.22 & 324.62 \\ \cline{2-6} 
			& $ N4 $ & \multicolumn{1}{c|}{128} & 5 & {31.68} & 238.46 \\ \hline
		\end{tabular}
	\end{adjustbox}
	\caption{Different versions of our EfficientNeRF with trade-off between synthesized accuracy and rendering speed. }
	\label{tab:trade-off}
\end{table}

% \begin{table}[h]
% 	\begin{adjustbox}{width=1.0 \columnwidth,center}
% 		\begin{tabular}{l|c|cc|c|c}
% 			\hline
% 			\multirow{2}{*}{} & \multirow{2}{*}{Version} & \multicolumn{2}{c|}{\# Sampling} & \multirow{2}{*}{PSNR ($\uparrow$)} & \multirow{2}{*}{{Rendering Speed (FPS, $\uparrow$)}} \\ \cline{3-4}
% 			&  & \multicolumn{1}{c|}{$ N_c $} & $ N_s $ &  &  \\ \hline
% 			\multirow{4}{*}{EfficientNeRF} & $ N1 $ & \multicolumn{1}{c|}{64} & 1 & 29.54 & {493.62} \\ \cline{2-6} 
% 			& $ N2 $ & \multicolumn{1}{c|}{64} & 2 & 30.49 & 403.28 \\ \cline{2-6} 
% 			& $ N3 $ & \multicolumn{1}{c|}{96} & 3 & 31.22 & 324.62 \\ \cline{2-6} 
% 			& $ N4 $ & \multicolumn{1}{c|}{128} & 4 & {31.68} & 238.46 \\ \hline
% 		\end{tabular}
% 	\end{adjustbox}
% 	\caption{Different versions of our EfficientNeRF with trade-off between synthesized accuracy and rendering speed. }
% 	\label{tab:trade-off}
% \end{table}
%psnr = [29.87, 30.49, 31.22, 31.54]
%[523, 447, 324, 238]

\subsection{Qualitative Comparison}
%\paragraph{Qualitatively Comparison}
% in Fig. \ref{fig:qualitatively}.

%\begin{figure}[t]
	%	\centering
	%	\includegraphics[width=1.0\linewidth]{figs/visualization_comparison.png}
	%	\caption{Qualitatively comparison with the state-of-the-art methods on the Realistic Synthetic dataset \cite{NeRF} and the Real Forward-Facing dataset \cite{NeRF,LLFF}. }
	%	\label{fig:qualitatively}
	%\end{figure}

\begin{figure*}[t]
	\centering
	\includegraphics[width=0.99 \linewidth]{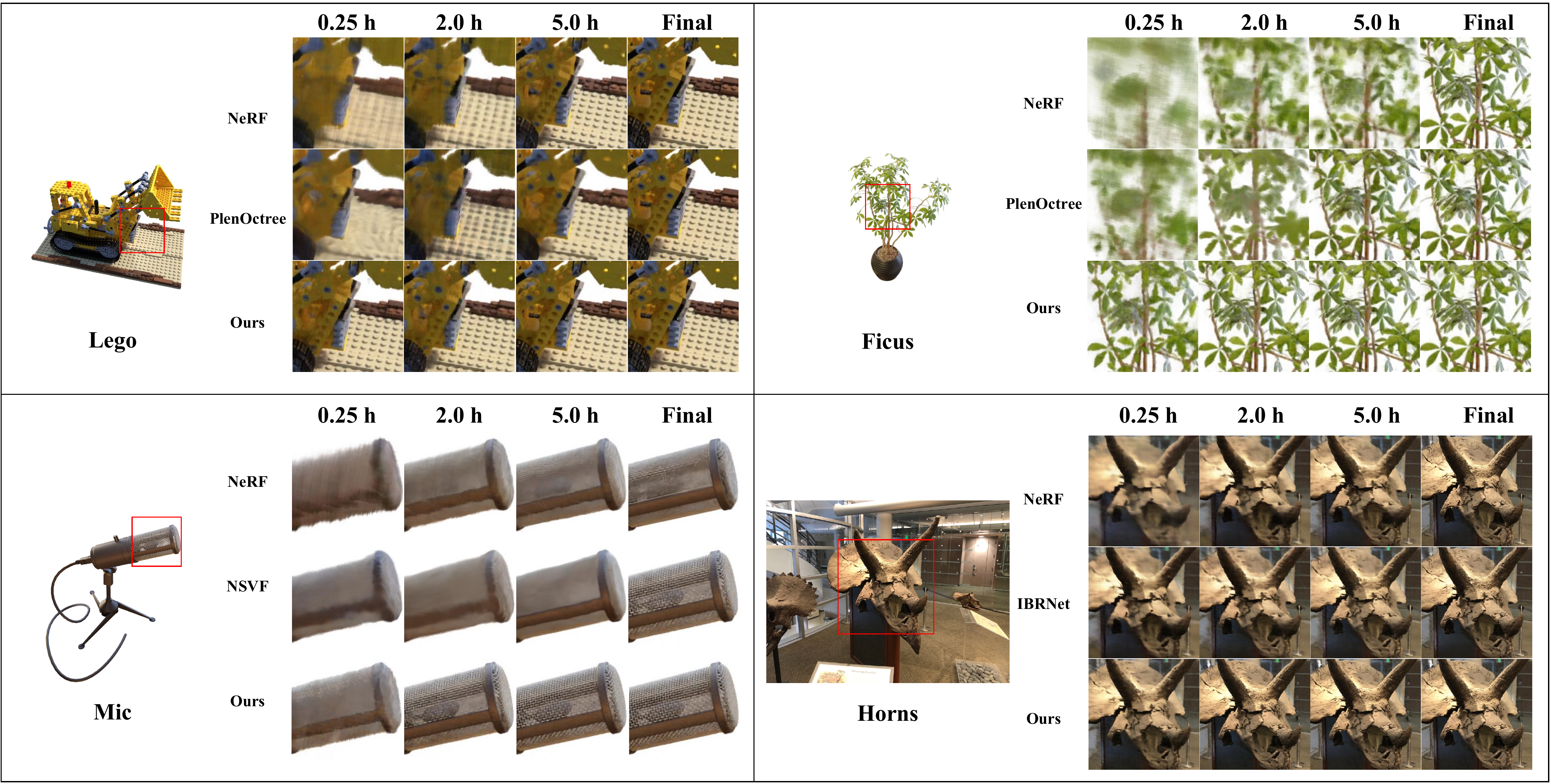}
	\caption{Qualitative comparison with state-of-the-art methods on the Realistic Synthetic dataset \cite{NeRF} and the Real Forward-Facing dataset \cite{NeRF,LLFF}. It is best viewed by zoom-in.}
	\label{fig:time_line}
\end{figure*}

We also demonstrate the performance of our method by visual comparison. As illustrated in Fig. \ref{fig:time_line}, we intuitively show the training visualization of different methods at $0.25$, 2, and 5 hours, and the final training time. In the timeline of training, ours already synthesizes detailed images within 1 hour, while other methods \cite{PlenOctree,NSVF,IBRNet} need to train 5 hours or longer to achieve similar performance. Also, compared with IBRNet \cite{IBRNet} based on image prior, our method is trained from scratch while outperforming it within $0.25$-hour training.

\subsection{Ablation Studies}
\paragraph{Networks}
We explore the influence of the different sizes of the coarse and fine networks, as shown in Table \ref{tab:mlp}.  The baseline combination is two identical standard MLPs, which come from the original NeRF \cite{NeRF}. The accuracy is the best under the longest running time. The combination of two lightweight MLPs yields opposite performance, which indicates the effect of standard MLPs. 

It is found that lightweight coarse MLP plus standard fine MLP yields nearly the same accuracy and fast rendering speed. It reveals that the size of fine MLP mainly determines the final synthesis quality.  We thus adopt the final combination as the network of our EfficientNeRF.

\begin{table}[t]
	\begin{adjustbox}{width=1.0\columnwidth,center}
		\begin{tabular}{c|c|c|c|c|c}
			\hline
			\multicolumn{2}{c|}{{Coarse MLP}}     & \multicolumn{2}{c|}{{Fine MLP}}   & \multicolumn{1}{c|}{\multirow{2}{*}{{\makecell{Time \\ (s / iter, $\downarrow$)}}}} & \multicolumn{1}{c}{\multirow{2}{*}{{PSNR($\uparrow$)}}} \\ \cline{1-4}
			\multicolumn{1}{c|}{{Lightweight}} & \multicolumn{1}{c|}{{Standard}} & \multicolumn{1}{c|}{{Lightweight}} & \multicolumn{1}{c|}{{Standard}} & \multicolumn{1}{c|}{}                                        & \multicolumn{1}{c}{}                               \\ \hline
			&   \checkmark  &      &    \checkmark  & 0.184  & \textbf{31.01}  \\ \hline
			\checkmark &      &   \checkmark   &     &   0.121    &  29.28  \\ \hline
			&   \checkmark   &   \checkmark   &     &    0.132   &   29.39       \\ \hline
			\checkmark&      &      &  \checkmark   &    \textbf{0.138}  &     \textbf{30.96}        \\ \hline
		\end{tabular}
	\end{adjustbox}
	\caption{Performance of different combinations between lightweight and standard MLPs at the coarse and fine stages of the original NeRF \cite{NeRF}. The batch size is 1024.}
	\label{tab:mlp}
\end{table}

% \vspace{-0.1in}
\paragraph{Efficient Modules}
As shown in Table \ref{tab:training_contribution}, we evaluate  performance of the proposed efficient modules. First, representing the color in different directions by Spherical Harmonic (SH) \cite{PlenOctree} is in favor of the performance. Second, our lightweight coarse MLP and coarse valid sampling accelerate the training process. Third, our fine pivotal sampling further reduces the training time and improves synthesis accuracy. It performs better than the original probabilistic sampling of original NeRF \cite{NeRF}.

\begin{table}[t]
	\begin{adjustbox}{width=1\columnwidth,center}
		\begin{tabular}{cccc}
			\hline
			Method                        & PSNR ($\uparrow$)           & Time ($\downarrow$)            & Improvement ($\uparrow$)       \\ \hline
			NeRF \cite{NeRF}                          & 31.01          & 0.184 s / iter      & -                  \\ 
			+ SH \cite{PlenOctree}                          & 31.57          & 0.183 s / iter      & -                  \\ \hline
			+ Lightweight Coarse MLP            & 31.52          & 0.137 s / iter     & 25.54\%  \\
			+ Coarse Valid  Sampling   & 31.49 & 0.085 s / iter   & 53.80\% \\
			+ Fine Pivotal Sampling & \textbf{31.68}     & \textbf{0.021 s / iter} & \textbf{88.58\%}  \\
			\hline
		\end{tabular}
	\end{adjustbox}
	\caption{Contributions of our proposed modules to the training time on the Realistic Sythetic dataset \cite{NeRF}.}
	\label{tab:training_contribution}
\end{table}

% \vspace{-0.1in}
\paragraph{Sampling Strategy}
We compare our efficient sampling with the common uniform sampling and original sampling of NeRF. The result is presented in Table \ref{tab:sampling}. First, uniform sampling is adopted by IBRNet \cite{IBRNet} and MVSNeRF \cite{MVSNeRF}. It is faster than NeRF sampling while achieving lower accuracy. Second, NeRF \cite{NeRF} and PlenOctree \cite{PlenOctree} adopt NeRF sampling and achieve high performance. However, the training time is very long. Finally,  our efficient sampling successfully accelerates all these methods while achieving comparable or even better accuracy.  

\begin{table}[t]
	\begin{adjustbox}{width=0.99\columnwidth,center}
		\begin{tabular}{c|c|c|c|c|c}
			\hline
			\multirow{2}{*}{Method}     & \multicolumn{3}{c|}{Sampling Strategy}                                                                              & \multirow{2}{*}{\makecell{Training Time\\ (Hours)}} & \multirow{2}{*}{PSNR} \\ \cline{2-4}
			& \multicolumn{1}{c|}{Uniform} & \multicolumn{1}{c|}{NeRF\cite{NeRF}} & \multicolumn{1}{c|}{{Ours}} &   &     \\ \hline
			\multirow{3}{*}{NeRF\cite{NeRF}} &    \checkmark    &  & & 41     & 30.06   \\ \cline{2-6}  &      &    \checkmark   & & 56  & 31.01  \\ \cline{2-6} 
			&   & &    \checkmark      &  \textbf{6}     & \textbf{31.25}        \\ \hline
			\multirow{2}{*}{IBRNet \cite{IBRNet}}  &  \checkmark  &  &  & 2  & 25.49 \\ \cline{2-6} 
			&   &  & \checkmark  & \textbf{1}   & \multicolumn{1}{l}{\textbf{26.23}}  \\ \hline
			\multirow{2}{*}{MVSNeRF \cite{MVSNeRF}}    &   \checkmark    &       &     & 0.25   & 27.21    \\ \cline{2-6} 
			&       &       &      \checkmark     &  \textbf{0.18}      &  \multicolumn{1}{l}{\textbf{28.03}}  \\ \hline
			\multirow{2}{*}{PlenOctree \cite{PlenOctree}} &     &    \checkmark  &        & 58      & \textbf{31.71}   \\ \cline{2-6} 
			&    &    &  \checkmark  & \textbf{6}    & \multicolumn{1}{l}{31.66}  \\ \hline
		\end{tabular}
	\end{adjustbox}
	\caption{Effect of our efficient sampling strategies when combined with different NeRF-based Methods.}
	\label{tab:sampling}
\end{table}

\begin{table}[t]
	\begin{adjustbox}{width=1.0\columnwidth,center}
		\begin{tabular}{c|c|c|c}
			\hline
			{\makecell{Scene \\ Representation}}  &  {Memory}    & {Caching Time}   & {Querying Time}  \\ \hline
			\makecell{ Dense Voxels}   &  16 GB &  16.55 ms & 13.64 ms \\ \hline
			\makecell{Sparse Tensor \\ (Minkowski Engine \cite{ME})} &  2.1 GB    & 24.72 s & 121.21 ms  \\ \hline
			\makecell{Octree\\ (PlenOctree \cite{PlenOctree})}               & 2.6 GB  &  14.51 s  & 18.84 ms   \\ \hline
			\textbf{\makecell{NerfTree \\ (Ours)}}   &  2.8 GB &   22.43 ms & 15.39 ms    \\ \hline
		\end{tabular}
	\end{adjustbox}
	\caption{Comparison of different data structures when caching and querying all points in a common 3D scene with size $1,024\times 1,024 \times 1,024$ and $20\%$ valid samples.}
	\label{tab:scene_representation}
\end{table}

% \vspace{-0.1in}
\paragraph{Scene Representation}
Before applying NerfTree, other data structures, such as dense voxels, sparse tensor \cite{ME}, and Octree \cite{PlenOctree}, can also be used to cache the trained scene for fast synthesis of novel views. To compare their efficiency, we calculate memory consumption and running time when storing and querying a whole scene with resolution $1,024 \times 1,024 \times 1,024$ and $20\%$ of valid space.  Each voxel has a 4D feature $(r,g,b,\sigma)$ with the same data type. 

The results are shown in Table \ref{tab:scene_representation},  Dense voxels representation spends the least time in caching and querying while requiring $16.0$ GB memory. The required memory by Sparse Tensor (Minkowski Engine \cite{ME}) is the smallest. But as it adopts hash structure making caching and query longer process. PlenOctree \cite{PlenOctree} balances memory consumption and query time. However, its caching time is long because of its internal optimization. 
Our NerfTree representation performs the best with all these three aspects. It does not require much storage memory, and cache and query 3D points at a very fast speed. 

\section{Conclusion}
\label{sec:con}
In this paper, we have presented Efficient Neural Radiance Fields (EfficientNeRF) to accomplish accurate representation of 3D scenes and synthesis of novel view images at a fast speed. We studied the distribution of density and weight and proposed valid sampling at the coarse stage and pivotal sampling at the fine stage. These two sampling strategies are efficiently handle the important samples, thus saving a great amount of computation. Also, we designed NerfTree for NeRF-based methods to cache 3D scenes. It yields faster speed than state-of-the-art methods \cite{NSVF,PlenOctree,KiloNeRF} during testing. 

\vspace{-0.1in}
\paragraph{Limitations and Future Work} 
Our EfficientNeRF achieves fast and accurate 3D scene representation and view synthesis. It still needs to train from scratch when handling novel scenes. This is also a common issue in other state-of-the-art NeRF-based methods \cite{PlenOctree}. Although we combined images prior with our efficient sampling in Table \ref{tab:sampling}, the synthesis accuracy is limited. In future work, we will improve generalization of EfficientNeRF and aim to achieve competitive accuracy when there is no finetuning in novel scenes. 

%%%%%%%%% REFERENCES
{\small
	\bibliographystyle{ieee_fullname}
	\bibliography{cvpr}
}

\end{document}